\documentclass{article}

\usepackage{microtype}
\usepackage{graphicx}
\usepackage{booktabs} 

\usepackage{hyperref}

\usepackage[accepted]{icml2023}

\usepackage{amsmath}
\usepackage{amssymb}
\usepackage{mathtools}
\usepackage{amsthm}

\usepackage[capitalize,noabbrev]{cleveref}

\theoremstyle{plain}

\theoremstyle{definition}

\theoremstyle{remark}

\usepackage[textsize=tiny]{todonotes}
\usepackage{multirow}
\usepackage{wrapfig}
\usepackage{colortbl}

\usepackage{enumitem}
\usepackage{microtype}
\usepackage{algpseudocode}
\usepackage{caption}
\usepackage{subcaption}
\usepackage{makecell}

\setlist[itemize]{align=parleft,left=0pt,topsep=1mm,itemsep=0mm,parsep=1mm}

\definecolor{azure(colorwheel)}{rgb}{0.0, 0.5, 1.0}
\definecolor{nicegreen}{rgb}{0.0, 0.7, 0.1}
\definecolor{CuGray}{gray}{0.9}
\definecolor{amethyst}{rgb}{0.6, 0.4, 0.8}
\definecolor{black}{rgb}{0.0, 0.0, 0.0}
\definecolor{steelblue}{rgb}{0.27, 0.51, 0.7}
\definecolor{brightcerulean}{rgb}{0.11, 0.67, 0.84}
\definecolor{postechred}{rgb}{0.784, 0.003, 0.313}

\newcolumntype{g}{>{\columncolor{CuGray}}c}
\newcolumntype{z}{>{\columncolor{CuGray}}l}

\renewcommand{\paragraph}[1]{\noindent\textbf{#1.}\,\,}

\usepackage{xspace}

\def\onedot{.\@\xspace}
\def\eg{\emph{e.g}\onedot} 
\def\ie{\emph{i.e}\onedot}

\newcommand{\Fref}[1]{Fig.~\ref{#1}}
\newcommand{\Tref}[1]{Table~\ref{#1}}

\newcommand{\calD}{{\mathcal{D}}}

\newcommand{\calR}{{\mathcal{R}}}
\newcommand{\calS}{{\mathcal{S}}}

\newcommand{\btheta}{\mbox{\boldmath $\theta$}}

\newcommand{\be}{\begin{eqnarray}}
\newcommand{\ee}{\end{eqnarray}}
\newcommand{\bee}{\begin{eqnarray*}}
\newcommand{\eee}{\end{eqnarray*}}

\newcommand{\matrixb}{\left[ \begin{array}}
\newcommand{\matrixe}{\end{array} \right]}   

\icmltitlerunning{Submission and Formatting Instructions for ICML 2023}

\begin{document}

\twocolumn[
\icmltitle{SYNAuG: Exploiting Synthetic Data for Data Imbalance Problems}



\icmlsetsymbol{equal}{*}

\begin{icmlauthorlist}

\icmlauthor{Moon Ye-Bin}{equal,yyy}
\icmlauthor{Nam Hyeon-Woo}{equal,yyy}
\icmlauthor{Wonseok Choi}{comp}
\icmlauthor{Nayeong Kim}{sch}
\icmlauthor{Suha Kwak}{sch,comp}
\icmlauthor{Tae-Hyun Oh}{yyy,comp,sch}
\end{icmlauthorlist}

\icmlaffiliation{yyy}{Dept. of Electrical Engineering, POSTECH, Pohang, South Korea}
\icmlaffiliation{comp}{Grad. School of Artificial Intelligence, POSTECH, Pohang, South Korea}
\icmlaffiliation{sch}{Dept. of Computer Science and Engineering, POSTECH, Pohang, South Korea}

\icmlcorrespondingauthor{Tae-Hyun Oh}{taehyun@postech.ac.kr}

\icmlkeywords{Machine Learning, ICML}

\vskip 0.3in
]



\printAffiliationsAndNotice{\icmlEqualContribution} 

\begin{abstract}
Data imbalance in training data often leads to biased predictions from trained models, which in turn causes ethical and social issues. A straightforward solution is to carefully curate training data, but given the enormous scale of modern neural networks, this is prohibitively labor-intensive and thus impractical. Inspired by recent developments in generative models, this paper explores the potential of synthetic data to address the data imbalance problem. To be specific, our method, dubbed SYNAuG, leverages synthetic data to equalize the unbalanced distribution of training data. Our experiments demonstrate that, although a domain gap between real and synthetic data exists, training with SYNAuG followed by fine-tuning with a few real samples allows to achieve impressive performance on diverse tasks with different data imbalance issues, surpassing existing task-specific methods for the same purpose.

\end{abstract}

\section{Introduction}

Modern machine learning resides in the over-parameterization regime~\citep{neyshabur2018the, pmlr-v97-allen-zhu19a}. Deep Neural Networks (DNNs) overfit the given abundant and diverse labeled data and show outstanding performance on the test data~\citep{deng2009imagenet, lin2014microsoft}. However, DNNs are vulnerable to biased features and distribution, \eg, shortcut learning~\citep{geirhos2020shortcut, NEURIPS2020_71e9c662, jacobsen2018excessive}, which is caused by the data distribution. Prior works have focused on the model architecture and algorithms in general~\citep{sambasivan2021everyone} to tackle this problem.

While the advancement of architectures and algorithms is undeniably crucial, the fundamental solution lies in prioritizing the quality and quantity of data.
In long-tail recognition~\citep{cui2019class}, we can enhance the accuracy of classes having a few number of data points by collecting more data for those classes.
It is important to acknowledge that collecting such data demands substantial resources in terms of human effort, time, and financial investment.

It leads us to ask the question, \textbf{``Can recent diffusion models be leveraged to address the data challenge?''} The success of generative diffusion models~\citep{rombach2022high} has been remarkable. These models demonstrate their ability to generate highly realistic images. Moreover, text-to-image diffusion models offer remarkable controllability, enabling users to generate images with short sentences. 
For example, we can generate images of waterbirds on the water  or land by changing the prompt to contain "on the water" or "on the land."
Likewise, diffusion models could be a plausible resource for the quality and quantity of data, which is the fundamental solution to data-caused issues.

In this work, our objective is to investigate the efficacy of synthetic data in addressing data-related challenges. For these experiments, we propose SYNAuG, the pipeline to augment the data and train the models. Initially, we address the class imbalance by leveraging text-to-image diffusion models to ensure uniformity in the number of samples across all classes. Subsequently, we apply off-the-shelf task-specific algorithms to train the classifier. Following training on this combined dataset comprising both original and synthetic data, we further optimize the last layer using uniformly sub-sampled original data.

Our investigation into the proposed SYNAuG method reveals several key findings. 
Firstly, while synthetic data generated by recent generative models can partially substitute real samples, a few real samples are crucial. We observe varying degrees of performance degradation when real samples are replaced with synthetic ones, highlighting the incompleteness of synthetic data as a perfect substitute for real-world data.
Secondly, we identify the importance of mitigating the domain gap between real and synthetic data. To address this issue, we propose several methods, including employing data augmentation techniques across real and synthetic data, and implementing fine-tuning strategies.
Lastly, despite the aforementioned concerns, SYNAuG addresses data imbalance issues such as long-tailed distributions, model fairness, and robustness to spurious correlation, showing potential for improving performance.
This offers valuable insights into the application of generative models for addressing real-world challenges.

\begin{figure}[t]
    \centering
    \includegraphics[width=1.0\linewidth]{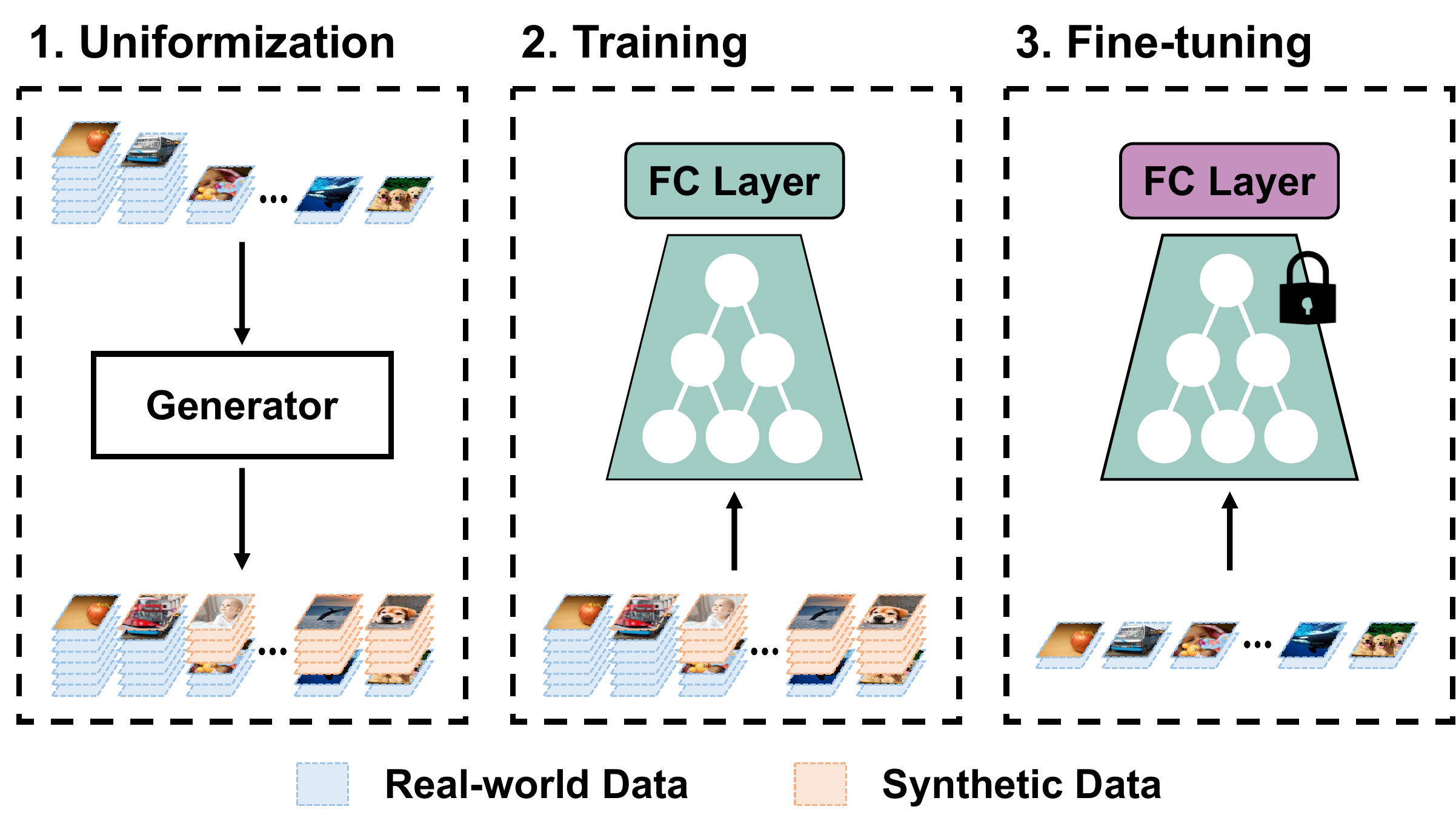}
    \caption{\textbf{Overview of SYNAuG process.}
    Given the imbalanced real-world data with the class labels, we first uniformize the imbalanced real data distribution by generating the synthetic samples that are conditioned on the class label.
    Second, we train a model with the uniformized training data.
    Finally, we fine-tune the last layer with the uniformly subsampled real-world data.}
    \label{fig:overview}
\end{figure}

\section{Background}

Data imbalance can lead to suboptimal generalization and many challenges in practical application scenarios, \eg, finance, healthcare, and autonomous driving.
The data imbalance problem is a common source of different imbalance sub-problems: long-tailed recognition, model fairness, and model robustness to spurious correlation.

\paragraph{Long-tailed recognition}
Long-tailed distribution is inherent to real-world~\citep{cui2019class,zhang2023deep}.
There are prior work in the realm of re-balancing classes, including re-sampling~\citep{shen2016relay,park2022majority} and re-weighting~\citep{samuel2021distributional, ren2020balanced, cui2019class},  
designing loss functions~\citep{ryou2019anchor, lin2017focal, ren2020balanced}, 
model selection~\citep{wang2021longtailed}, 
and language-guided feature augmentation~\citep{yebin2023textmania}.

\paragraph{Model fairness} 
In fairness~\citep{narayanan2018translation, hardt2016equality, 10.1145/2783258.2783311}, researchers have tackled the issue of model bias, where accuracy varies based on sensitive attributes such as race, age, and ethnicity.
Model fairness is also related to data imbalance because the number of samples of some sensitive groups is lower than that of the major groups.
Similar to long-tailed recognition, fairness has predominantly been tackled using loss weighting~\citep{jung2023reweighting} and batch sampling~\citep{kamiran2012data, roh2020fairbatch}.

\paragraph{Spurious correlation}
The spurious correlation problem is related to the robustness of models against misleading correlations.
DNNs are susceptible to falling into shortcuts that capture the most frequently observed patterns in a class regardless of true causality;
it is called spurious correlation or shortcut problems~\citep{geirhos2020shortcut,  kirichenko2023last}.
The spurious correlation problem is also dealt with similar approaches to the above two tasks: reweighting~\citep{sagawa2019distributionally,kim2022learning}, sampling~\citep{idrissi2022simple, sagawa2020investigation}, and post-calibration~\citep{liu2021just, kirichenko2023last}.

\paragraph{Summary of data imbalance problems}
While researchers have developed algorithms for each task separately, three different tasks primarily arising from data imbalance have been tackled in the shared perspective, \ie, up-weight loss values or sampling probabilities of minor groups using group or sensitive information.
However, most have focused more on algorithmic parts; thus, the inherent imbalance may still remain.
In this work, we shed light on the overlooked convention to go beyond the given bounded dataset.
We exploit the synthetic data from the generative foundation models~\citep{rombach2022high, saharia2022photorealistic, nichol2021glide} to take flexibility and controllability.

\paragraph{Using synthetic data in machine learning tasks}
Synthetic data is beneficial to overcome the lack of data and sensitive issues of data, \eg, licensing and privacy concerns.
Several approaches~\citep{jahanian2021generative, oh2018learning, zhang2021datasetgan} have started to leverage synthetic data for their tasks of interest, but do not consider imbalance problems.
Recently, deep generative models~\citep{rombach2022high, saharia2022photorealistic, nichol2021glide} have shown promising results in generating realistic and high-quality samples, stemming from the goal of modeling the real data distribution.
In this work, we explore the use of a pre-trained foundation diffusion model~\citep{rombach2022high} to observe whether synthetic data is effective in mitigating data imbalance problems.

\section{Exploiting Synthetic Data}
We propose SYNAuG, which leverages synthetic data to mitigate data imbalance problems from a data perspective.
As illustrated in \Fref{fig:overview}, we first uniformize the imbalanced data by generating synthetic data, then train the model on the uniformized data, and finally fine-tune the last layer with a few original data uniformly subsampled from each class.
We exploit a recent powerful generative model, Stable Diffusion~\citep{rombach2022high}, to generate synthetic data of corresponding classes or attributes with the controllable prompt.
Since they are trained on a large number of web data, it would be considered to cover and model a wide distribution of the real world.
Exploiting these favorable properties, we generate supporting data to alleviate the imbalance of the data distribution.

While uniformizing with synthetic data is simple and effective, there is still room to improve its performance because of the domain gap.
To improve further by mitigating the domain gap, we propose utilizing two simple methods.
First, we propose to leverage Mixup~\citep{zhang2017mixup} during training to augment samples to be interpolated samples between real and synthetic samples, \ie, domain Mixup.
Second, we propose to fine-tune a classifier on the subsampled uniform original data from the original training data after the first training stage.
The fine-tuned classifier would lead to more accurate recognition of target data. 
Algorithm \ref{alg:ours} is a pseudo code of SYNAuG.

\begin{algorithm}[t]
\caption{SYNAuG}\label{alg:ours}
    \begin{algorithmic}
        \Require Real data $\calR$, Generator $G$, image encoder $f$ and linear classifier $g$ parameterized with $\btheta_f$ and $\btheta_g$, and training algorithms $\texttt{Alg}^{M}(\cdot)$ and $\texttt{Alg}(\cdot)$ with and without Mixup, respectively. \\
        \textbf{1.} Training data: $\calD = \texttt{Uniformization}(\calR, \calS)$, where $\calS$ is synthetic data generated from $G$. \\
        \textbf{2.} Training $f$ and $g$ : $f, g = \texttt{Alg}^{M}(f, g, \calD)$  \Comment{Both $\btheta_f$ and $\btheta_g$ are updated} \\
        \textbf{3.} Fine-tuning $g$ : $g = \texttt{Alg}(g, \calR)$ \Comment{Last layer is fine-tuned, \ie, only $\btheta_g$ is updated, with the subset of real data, which of distribution is uniform.}
    \end{algorithmic}
\end{algorithm}

\paragraph{Details of image generation and uniformization}
We can produce high-quality synthetic images across various classes by employing Stable Diffusion~\citep{rombach2022high}.
We can also achieve data controllability based on a text prompt containing a class name to represent class semantics.  
The synthetic data is utilized to populate and uniformize each training dataset, including CIFAR100-LT~\citep{cao2019learning}, ImageNet100-LT~\citep{jiang2021self}, UTKFace~\citep{zhifei2017cvpr}, and Waterbirds~\citep{sagawa2019distributionally}.

To generate synthetic images for CIFAR100-LT, we use the prompt ``a photo of \{\texttt{modifier}\} \{\texttt{class}\},'' leveraging a set of 20 modifiers recommended by ChatGPT~\citep{ouyang2022training} for each class.
ImageNet100-LT has 100 classes, and the image resolution is 256$\times$256, ranging from a maximum of 1,280 to a minimum of 6 images per class.
However, merely 3 classes possess over 500 samples per class. Consequently, we have added synthetic data to ensure 500 samples for the remaining classes.
We input the prompt ``a photo of \{\texttt{class}\}, ...'' during the generation process.

UTKFace is composed of 23,708 images with age, gender, and race labels.
Age is from 0 to 116; 
gender is male or female;
races are White, Black, Asian, Indian, and others.
We use race annotation as group labels, serving as sensitive attributes, except for others. 
Additionally, gender annotation functions as class labels.
For generating synthetic data, we provide a prompt incorporating race, gender, and age, structured as ``Portrait face photo of \{\texttt{race}\} \{\texttt{gender}\}, \{\texttt{age}\} year old ...''.

Waterbirds is synthesized 
using the CUB dataset~\citep{WelinderEtal2010} and the backgrounds.
The seabirds and waterfowl are grouped as waterbirds, while the other bird types are categorized as landbirds.
For generating synthetic data, we provide the prompt composed of water or land, such as ``The photo of water bird in \texttt{land} ...''.

\section{Experiments}
We evaluate our method for three sub-tasks: long-tailed recognition, model fairness, and model robustness to spurious correlation.
We provide key findings and the effectiveness of SYNAuG for data imbalance problems.

\subsection{Experimental Settings}
\paragraph{Long-tailed recognition}
We employ two long-tail datasets: CIFAR100-LT~\citep{cao2019learning} and ImageNet100-LT~\citep{jiang2021self}.
The test sets for them are the same as the original one.
The classes in the long-tailed datasets are divided into three groups: Many-shot (more than 100 samples), Medium-shot (20-100 samples), and Few-shot (less than 20 samples).
The imbalance factor (IF) corresponds to the skewness of the training data; high IF means high skewness and a more challenging setting.
We evaluate under the standard IFs of 100, 50, and 10, following the prior work~\citep{alshammari2022long}.
We use ResNet32 for CIFAR100-LT and ResNet50 for ImageNet100-LT.

\paragraph{Model fairness}
We employ UTKFace~\citep{zhifei2017cvpr} composed of 23,708 images with age, gender, and race labels.
We use race annotation as a sensitive attribute (group label) and gender as the class label.
We use fairness metrics, which have been proposed to measure the fairness performance of models:
Demographic Parity (DP)~\citep{10.1145/2783258.2783311},
Equal Opportunity (EO)~\citep{hardt2016equality, jung2022learning},
and Equalized Odds (ED)~\citep{hardt2016equality}.
These metrics are based on the difference in the performance of the learned classifiers depending on groups, \ie, the sensitive attributes.
Lower values of fairness metrics indicate that the model is fairer.

\paragraph{Model robustness to spurious correlation}
We use the Waterbirds dataset~\citep{sagawa2019distributionally}, which is a synthetic dataset created by combining images of birds from the CUB dataset~\citep{WelinderEtal2010} with backgrounds.
The birds are grouped into two categories: waterbirds, which include seabirds and waterfowl, and landbirds.
The Land and water background are spuriously correlated to landbirds and waterbirds, respectively.
We report the result over 5 independent runs using the code from DFR~\citep{kirichenko2023last}.
We reproduce the BaseModel and DFR and report the performance.\footnote{\url{https://github.com/PolinaKirichenko/deep_feature_reweighting}}

\subsection{Experimental Results}
\begin{figure}[t]
    \centering
    \begin{subfigure}[b]{0.7\linewidth}
        \centering
        \includegraphics[width=1.0\linewidth]{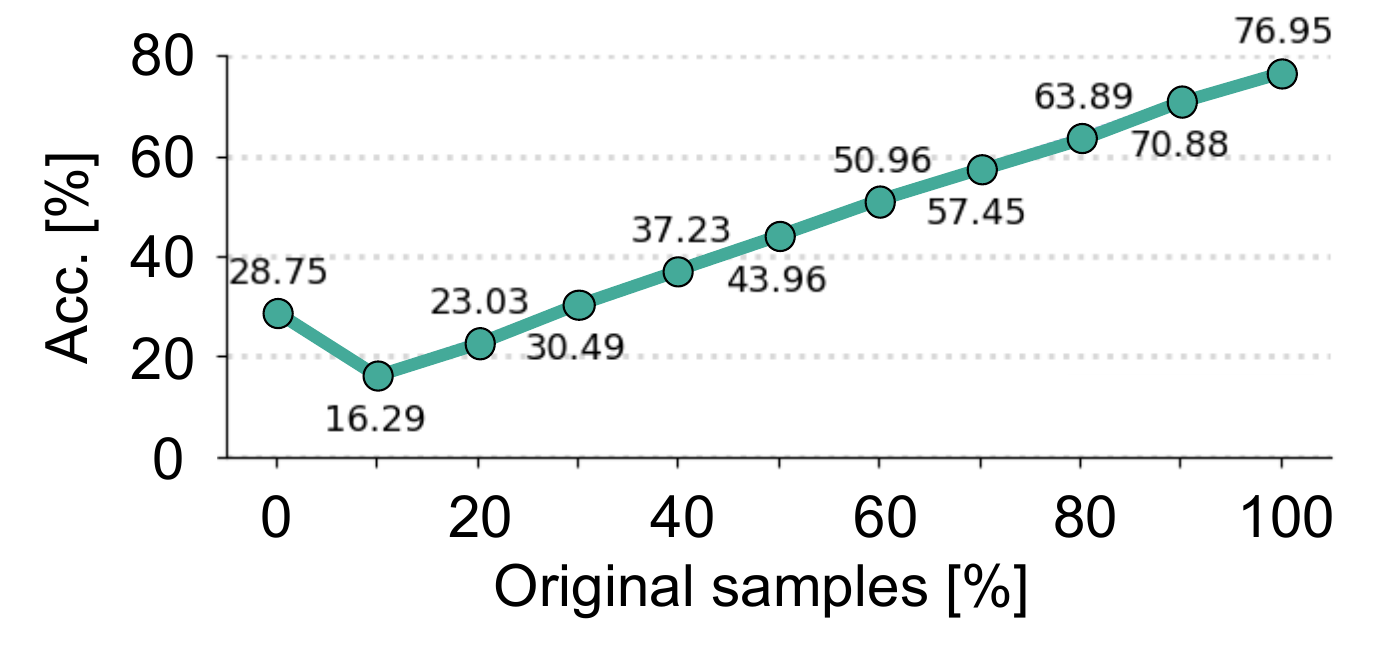}
        \caption{Class-wise replacement}
        \label{fig:classwise}
    \end{subfigure}
    \centering
    \begin{subfigure}[b]{0.7\linewidth}
        \centering
        \includegraphics[width=1.0\linewidth]{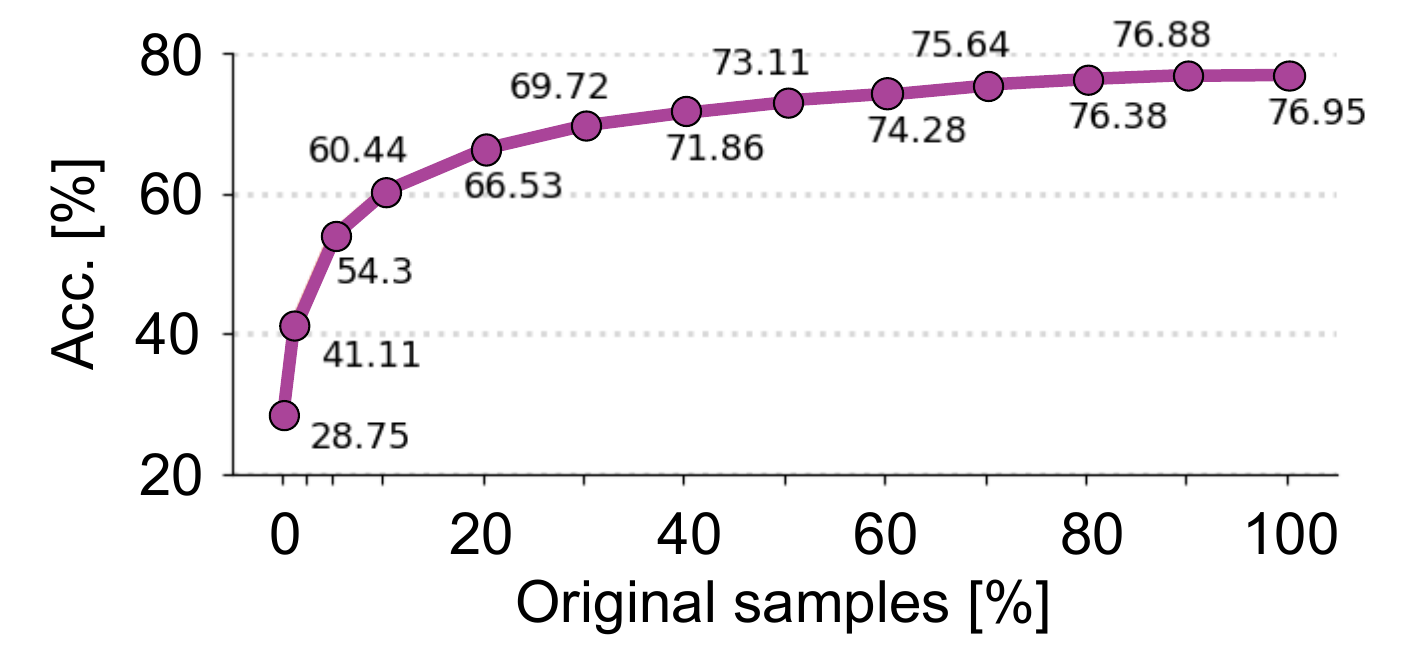}
        \caption{Instance-wise replacement}
        \label{fig:instwise}
    \end{subfigure}    
    \caption{\textbf{Replacement test.} 
    To investigate the effect on model performance when using original and synthetic data together, we replace the original data with synthetic ones in two ways: (a) class-wise and (b) the same ratio of instances across all classes.
    We use CIFAR100, which has 500 samples per class and 100 classes.
    }
    \label{fig:abl_ratio}
\end{figure}

\begin{figure}[t]
    \centering
        \begin{subtable}[c]{0.3\linewidth}
        \centering
        \resizebox{1.0\linewidth}{!}{
            \begin{tabular}{cc}
                \toprule
                & \textbf{Accuracy} \\ 
                \midrule
                Real  & 77.76 \\
                Syn.  & 70.56 \\
                \cmidrule{1-2}
                Total & 74.16 \\
                \bottomrule 
            \end{tabular}
        }
        \caption{Binary domain classification}    
        \label{fig:domain_cls}
    \end{subtable}
    \hspace{8mm}
    \centering
        \begin{subfigure}[c]{0.45\linewidth}
        \centering
        \includegraphics[width=1.0\linewidth]{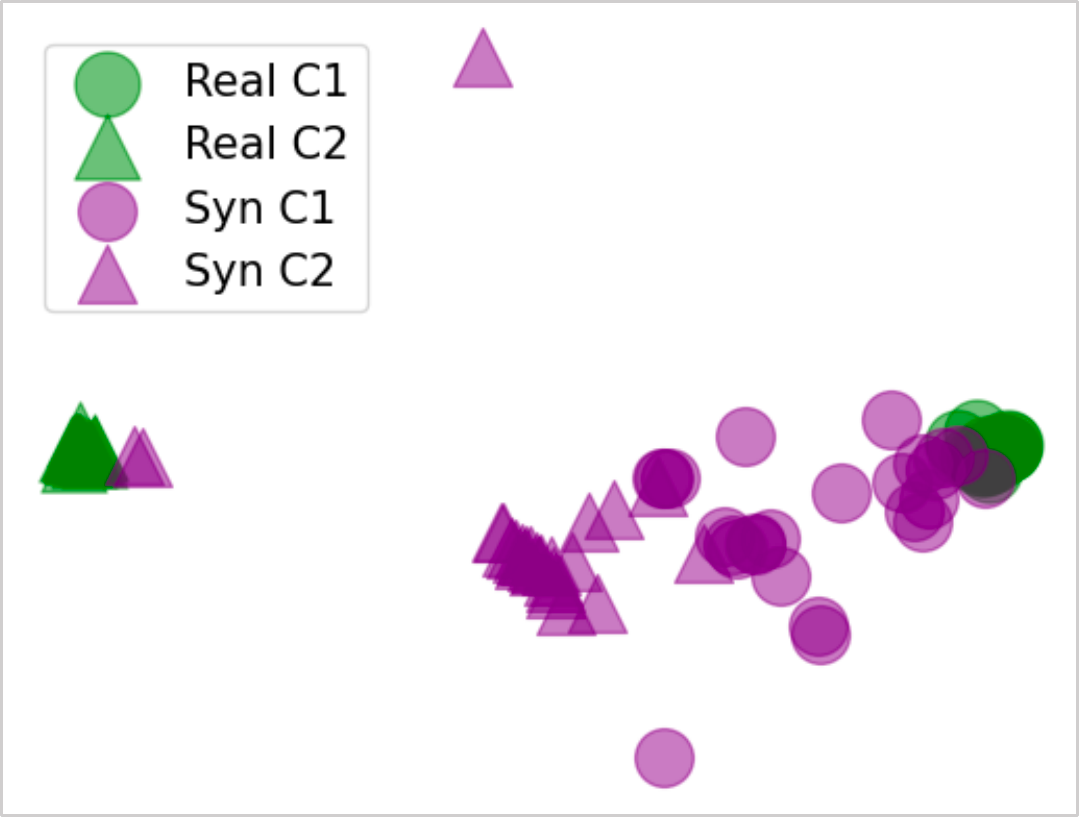}
        \caption{Feature visualization}
        \label{fig:domain_feat}
    \end{subfigure}    
    \caption{\textbf{Domain gap between real and synthetic data.} 
    We test domain gap empirically with (a) binary domain classification and (b) feature visualization.
    For classification, we use 2.5k samples for each real and synthetic domain and train only one fully-connected layer on the features extracted from pre-trained model.
    For visualization, the features are extracted from the pre-trained model on CIFAR100.
    C1 and C2 denote different classes.
    }\vspace{2mm}
    \label{fig:domain}
\end{figure}

\noindent\textbf{Is synthetic data effective? Yes but not equal to real data.}
In \Tref{fig:abl_ratio}, we check the effectiveness of synthetic data compared to real data by replacing some of the original data with synthetic data in two ways of settings.
The original dataset has the same amount of data in each class.
In the \textbf{first setting}, we completely replace original data belonging to specific classes with synthetic data. Note that certain classes have no real sample but only synthetic samples. 
In the \textbf{second setting}, we uniformly replace the original data with synthetic data, which means all classes have the same amount of original and synthetic data.

To isolate the effects of synthetic data, we do not apply Mixup and last-layer fine-tuning.
The results reveal contrasting patterns between the two settings. In the first setting, we observe a linear decline in performance as the number of classes lacking original data increases (See \Fref{fig:classwise}).
Conversely, the second setting shows a logarithmic decrease in performance as more original data are uniformly replaced with synthetic data (See \Fref{fig:instwise}). 
Remarkably, using only 1\% of real data in the second setting yields a performance of 41.11\%, which is comparable to the 43.96\% achieved using 50\% real data in the first setting. 
It suggests that synthetic data help improve performance even with a small amount of original data, but still including at least a small amount of original data is crucial.
It may imply a complementary domain gap between synthetic and real data.

\noindent\textbf{Is there a domain gap between synthetic and real data? Yes.}
The above experiments also show the domain gap may still exist even with high-quality synthetic data. To check the presence of a domain gap, we conduct domain classification and visualization of the features from both real and synthetic data (See \Fref{fig:domain}). As shown in \Fref{fig:domain_cls}, the domain classification performance is 74.16\%. Considering that 50\% means no domain gap, the result implies that 
there is an apparent domain difference classifiable. 
As shown in \Fref{fig:domain_feat}, the features of Syn C2 are closer to Syn C1 than Real C2. This observation provides empirical evidence of a domain gap existing between real and synthetic data.

\begin{table}[t]
\centering
\caption{\textbf{Method to reduce domain gap.} 
    We use CIFAR100-LT.
    Each component, Modifier, Mixup, Re-train, and Finetune, means we use the class-related modifiers in the prompt, use Mixup augmentation during training, and re-train or finetune the last layer after training, respectively.
(e) stands for our SYNAuG.}
    \label{tab:ablation}
\resizebox{1.0\linewidth}{!}{
    \begin{tabular}{cccccccc}
        \toprule
        \multirow{2}[2]{*}{} & \multirow{2}[2]{*}{\textbf{Modifier}} & \multirow{2}[2]{*}{\textbf{Mixup}} & \multirow{2}[2]{*}{\textbf{Re-train}} & \multirow{2}[2]{*}{\textbf{Finetune}} & \multicolumn{3}{c}{\textbf{IF}} \\ \cmidrule{6-8}
        & & & & & \textbf{100} & \textbf{{50}} & \textbf{{10}} \\
        \midrule
        (a) & & & & & 52.41 & 56.99 & 66.34 \\
        (b) & \checkmark & & & & 53.54 & 57.09 & 66.66 \\
        (c) & \checkmark & \checkmark & & & 55.45 & 58.69 & 66.84 \\
        (d) & \checkmark & \checkmark & \checkmark & & 57.31 & 60.34 & 67.90 \\
        (e) & \checkmark & \checkmark & & \checkmark & \textbf{58.59} & \textbf{61.36} & \textbf{69.01} \\
        \bottomrule 
    \end{tabular}
    }
\end{table}

\noindent\textbf{Do we need to reduce the domain gap? Yes.}
We propose several methods to reduce the domain gap between synthetic and real data:
Modifiers generate diverse synthetic samples, Mixup interpolates between synthetic and real samples, whereby the domain gap is mitigated by bridging two different domain data, and we fine-tune the last layer to get an appropriate hyperplane for real samples.
\Tref{tab:ablation} shows the effectiveness of our methods.
When we use modifiers in the prompt, we observe the performance gain (See (a) and (b) in \Tref{tab:ablation}).
We can achieve further improvement by utilizing Mixup (See (c) in \Tref{tab:ablation}).
As shown in \Tref{tab:ablation}-(d,e), we can achieve an additional improvement by adjusting the classifier towards the targeted real data and found that fine-tuning is more effective than re-training.

\begin{figure}[t]
    \begin{subfigure}[c]{0.9\linewidth}
        \centering
        \includegraphics[width=1.0\linewidth]{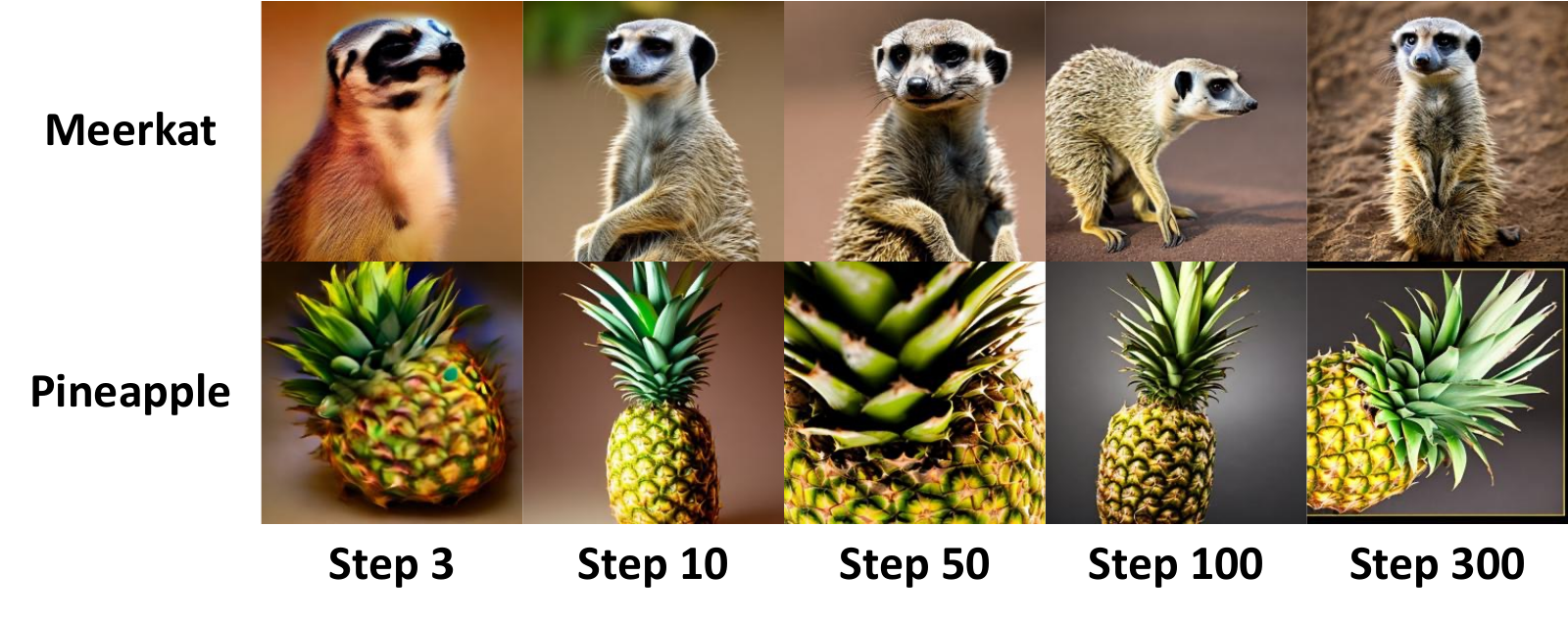}
    \end{subfigure}
    \centering
    \begin{subtable}[c]{0.8\linewidth}
    \resizebox{1.0\linewidth}{!}{
        \begin{tabular}{@{\,\,\,}lccccc}
            \toprule
            \textbf{Method}  & \textbf{\# step} & \textbf{Many} & \textbf{Medium} & \textbf{Few} & \textbf{All} \\
            \midrule
            CE$_\texttt{ CVPR'19}$   & & \textbf{61.85} & 15.83 &  0.29 & 32.06 \\
            \cmidrule{1-6}
            \multirow{5}{*}{SYNAuG}
            & 3    & 48.23 & 46.29 & 40.07 & 45.10 \\
            & 10   & 53.89 & \textbf{49.49} & 43.87 & \textbf{49.34} \\
            & 50   & 52.91 & 48.63 & \textbf{45.27} & 49.12 \\
            & 100  & 53.03 & 49.20 & 44.47 & 49.12 \\
            & 300  & 54.11 & 47.71 & 44.73 & 49.06 \\
            \bottomrule 
        \end{tabular}
        }
    \end{subtable}
\caption{\textbf{Ablation study according to sample quality.}
\textbf{(Top)} quality of the generated samples according to the number of steps, \textbf{(Bottom)} long-tailed recognition performance (\%) according to the different times of steps for generating synthetic data, which affects sample quality.
We use ImageNet100-LT with ResNet50.
}
\label{fig:imagenet100_lt}
\end{figure}

\noindent\textbf{Is more realistic synthetic data more effective? Not really.}
In \Fref{fig:imagenet100_lt}, we evaluate SYNAuG on ImageNet100-LT.
We conduct an ablation study to investigate the impact of data quality of SYNAuG by controlling the diffusion step parameter of Stable Diffusion~\citep{rombach2022high}, which is known to affect the quality of generated images.
As shown in \Fref{fig:imagenet100_lt}-(Top), generation quality is low with a minimal number of steps, but visually, the differences become less noticeable as the step count increases to a certain level.
Figure~\ref{fig:imagenet100_lt}-(Bottom) shows the quantitative results.
Compared to the CE method trained on the original long-tailed data, while the accuracy of the Many class is degraded, we achieve a large improvement in the Medium, Few, and even All cases regardless of the synthetic image quality.
However, there is a certain level of quality that exhibits a surge point in performance.
The quality difference becomes negligible when the step value exceeds a certain threshold.
This may hint that there exists a favorable gap that can complement real data.

\begin{table}[t]
\centering
\caption{\textbf{Long-tailed recognition performance on CIFAR100-LT.}
    We compare our SYNAuG with recent works.
    We report the Top-1 accuracy (\%) with different imbalance factors, \ie, IF=\{100, 50, 10\}.
    }
    \label{tab:cifar100_lt}
\resizebox{1.0\linewidth}{!}{
    \begin{tabular}{@{}lcccccc}
        \toprule
        \multirow{2}[2]{*}{\textbf{{Method}}} & \multicolumn{4}{c}{\textbf{IF=100}} & \multirow{2}[2]{*}{\textbf{{50}}} & \multirow{2}[2]{*}{\textbf{{10}}} \\
        \cmidrule{2-5}
        & \textbf{Many} & \textbf{Medium} & \textbf{Few} & \textbf{All} & & \\
        \midrule
        CE$_\texttt{ CVPR'19}$                       & 68.31 & 36.88 & 4.87 & 37.96 & 43.54 & 59.50 \\
        \cmidrule{1-7}
        SSD$_\texttt{ ICCV'21}$                          & - & - & - & 46.0 & 50.5 & 62.3 \\
        PaCo$_\texttt{ ICCV'21}$                  & - & - & - & 52.0 & 56.0 & 64.2 \\
        RISDA$_\texttt{ AAAI'21}$                   & - & - & - & 50.16 & 53.84 & 62.38 \\
        CE + CMO$_\texttt{ CVPR'22}$            & 70.4 & 42.5 & 14.4 & 43.9 & 48.3 & 59.5 \\
        LDAM + CMO$_\texttt{ CVPR'22}$            & 61.5 & 48.6 & 28.8 & 47.2 & 51.7 & 58.4 \\
        RIDE (3 experts) + CMO$_\texttt{ CVPR'22}$ & - & - & - & 50.0 & 53.0 & 60.2 \\
        Weight Balancing$_\texttt{ CVPR'22}$     & 72.60 & 51.86 & 32.63 & 53.35 & 57.71 & 68.67 \\
        \cmidrule{1-7}
        SYNAuG                                    & \textbf{74.06} & \textbf{56.63} & \textbf{42.83} & \textbf{58.59} & \textbf{61.36} & \textbf{69.01} \\
        \bottomrule 
    \end{tabular}
    }
\end{table}

\begin{table}[t]
    \centering
    \caption{\textbf{Fairness performance.} 
   (a) accuracy ($\uparrow$) and model fairness ($\downarrow$), 
   (b) compatibility with fairness algorithms, Group-DRO and Re-Sampling (RS), 
   (c) ablation study with data augmentation, Mixup and Cuxmix, 
   and (d) ablation study using the prior about sensitive attributes.}
   \label{tab:temps}
    \begin{subtable}{0.75\linewidth}
        \centering
        \resizebox{\linewidth}{!}{
            \begin{tabular}{lcccc}
            \toprule
            \textbf{Method} & \textbf{Accuracy} & \textbf{DP} & \textbf{ED} & \textbf{EO}\\
            \midrule
            \multicolumn{5}{c}{ResNet18} \\
            \midrule
            ERM & 93.9 & 0.0817 & 0.0632 & 0.0779\\
            SYNAuG & \textbf{94.1} & \textbf{0.0600} & \textbf{0.0462} & \textbf{0.0434}\\
            \midrule
            \multicolumn{5}{c}{ResNet50} \\
            \midrule
            ERM & 94.0 & 0.07272 & 0.05850 & 0.07269\\
            SYNAuG & \textbf{94.4} & \textbf{0.05432} & \textbf{0.03936} & \textbf{0.05472}\\
            \bottomrule
           \end{tabular}
        }
    \caption{\textbf{Fairness performance.}}
    \vspace{2mm}
    \label{tab:fairness_1}
    \end{subtable}

    \begin{subtable}{0.75\linewidth}
        \centering
        \resizebox{\linewidth}{!}{
            \begin{tabular}{lcccc}
            \toprule
            \textbf{Method} & \textbf{Accuracy} & \textbf{DP} & \textbf{ED} & \textbf{EO}\\
            \midrule
            ERM & 93.9 & 0.08174 & 0.06320 & 0.07790\\
            + Group-DRO & 93.9 & 0.07266 & 0.05819 & 0.06954\\
            + RS & 93.8 & \textbf{0.06360} & \textbf{0.04663} & \textbf{0.05808} \\
            \midrule
            SYNAuG & \textbf{94.0} & 0.06995 & 0.05630 & 0.06825\\
            + Group-DRO & 93.9 & 0.07110 & 0.05433 & 0.07045\\
            + RS & 93.6 & 0.06786  & 0.06439 & 0.08120 \\
            \bottomrule
           \end{tabular}
        }
       \caption{\textbf{Ablation with other algorithms}}
       \vspace{2mm}
       \label{tab:fairness_2}
   \end{subtable}

    \begin{subtable}{0.75\linewidth}
        \centering
        \resizebox{\linewidth}{!}{
            \begin{tabular}{lcccc}
            \toprule
            \textbf{Method} & \textbf{Accuracy} & \textbf{DP} & \textbf{ED} & \textbf{EO}\\
            \midrule
            ERM & 93.9 & 0.08174 & 0.06320 & 0.07790\\
            + Mixup & 93.9 & 0.07266 & 0.05819 & 0.06954\\
            + CutMix & 94.7 & 0.08265 & 0.06245 & 0.08333\\
            \midrule
            SYNAuG& 94.0 & 0.06995 & 0.05630 & 0.06825\\
            + Mixup & 94.2 & \textbf{0.06400} & \textbf{0.03793} & \textbf{0.04658}\\
            + CutMix & \textbf{94.9} & 0.07430 & 0.04745 & 0.06319\\
            \bottomrule
           \end{tabular}
        }
       \caption{\textbf{Augmentation ablation}}
       \vspace{2mm}
       \label{tab:fairness_3}
   \end{subtable}

    \begin{subtable}{0.75\linewidth}
        \centering
        \resizebox{\linewidth}{!}{
            \begin{tabular}{lcccc}
            \toprule
            \textbf{Method} & \textbf{Accuracy} & \textbf{DP} & \textbf{ED} & \textbf{EO}\\
            \midrule
            SYNAuG$^*$ & \textbf{94.0} & 0.07342 & 0.05805 & 0.07253\\ 
            SYNAuG & \textbf{94.0} & \textbf{0.06995} & \textbf{0.05630} & \textbf{0.06825}\\ 
            \bottomrule
            \addlinespace[0.2mm]
            \multicolumn{5}{l}{*Not use the sensitivity attribute}
           \end{tabular}
        }
       \caption{\textbf{Sensitivity ablation}}
       \label{tab:fairness_4}
   \end{subtable}
\end{table}

\begin{table}[t]
    \centering
           \caption{\textbf{Robustness to spurious correlation on Waterbirds.} 
       SYNAuG outperforms DFR consistently in worst-group accuracy.}
       \label{tab:robustness}
        \resizebox{0.63\linewidth}{!}{
            \begin{tabular}{lcc}
            \toprule
            \multirow{2}[2]{*}{\textbf{Method}} & \multicolumn{2}{c}{\textbf{Waterbirds}}\\
            \cmidrule{2-3}
            & \textbf{Worst} & \textbf{Mean}\\
            \midrule
            BaseModel & 73.7$_{\pm 3.04}$ & 90.4$_{\pm 0.21}$\\
            SYNAuG & 79.9$_{\pm 2.22}$ & 91.5$_{\pm 0.98}$\\
            \midrule
            DFR$^{\text{Tr}}_{\text{Val}}$$_\texttt{ ICLR'23}$ & 91.2$_{\pm 1.92}$ & 93.1$_{\pm 0.91}$\\
            SYNAuG (re-train) & 92.9$_{\pm 0.45}$ & 93.6$_{\pm 0.35}$\\
            SYNAuG (fine-tuning) & 93.2$_{\pm 0.42}$ & 94.8$_{\pm 0.11}$\\
            \bottomrule
           \end{tabular}}
\end{table}

\noindent\textbf{Does synthetic data help with data imbalance problems? Indeed.}
For the long-tailed recognition problem, there are significant improvements when we use synthetic data, as shown in \Tref{tab:cifar100_lt}.
When we compare the accuracy between CE and SYNAuG, performance is improved regardless of the skewness of the training data.
CE stands for the method trained with the Cross-Entropy loss on the original data.
The model performance is also improved on the fairness problem in \Tref{tab:fairness_1} and the spurious correlation problem in \Tref{tab:robustness}.
Our results demonstrate that synthetic data is helpful for data imbalance problems.

\noindent\textbf{Is synthetic data more helpful than the state-of-the-art algorithmic methods? Surprisingly, yes.}
In \Tref{tab:cifar100_lt}, we compare with recent prior arts: SSD~\citep{li2021self} and PaCo~\citep{cui2021parametric} for self-supervised learning, RISDA~\citep{chen2022imagine} and CMO~\citep{park2022majority} for data augmentation, and Weight Balancing~\citep{alshammari2022long} for the rebalance classifier.

In \Tref{tab:cifar100_lt}, our method outperforms most of the computing methods.
This is a stunning result in that it suggests that relieving the imbalance from the data point of view is simple but more effective than the conventional complex algorithmic methods.

\noindent\textbf{Is uniformization with synthetic data better than other data augmentation methods?}
In \Tref{tab:cifar100_lt_baseline}, compared to the popular image augmentation methods, Mixup~\citep{zhang2017mixup} and Cutmix~\citep{yun2019cutmix}, our performance is much higher.
The results are consistent with the one na\"ively applying Mixup to imbalanced data cases is known to be detrimental~\citep{yebin2023textmania}; thus, we distinctively apply Mixup after uniformizing data distribution, which makes a noticeable difference.

In \Tref{tab:fairness_3}, we evaluate the effect of data augmentations when we apply Mixup and Cutmix after uniformization. In this ablation study, we do not apply fine-tuning for clear comparison. Both augmentations improve the accuracy of ERM; Mixup also works in the fairness metrics. Compared to ERM with Mixup, SYNAuG shows higher accuracy and better fairness metrics. SYNAuG with Mixup outperforms more in accuracy and fairness metrics compared to ERM with Mixup.
Although Mixup or Cutmix is applied regardless of the data type, real or synthetic, it allows domain gap reduction and performance improvement.

\begin{table}[t]
\centering
    \caption{\textbf{Comparison with the baselines.}
    We use CIFAR100-LT.
    The second column denotes the data type used in uniformization.
    }
    \label{tab:cifar100_lt_baseline}
\resizebox{0.88\linewidth}{!}{
    \begin{tabular}{@{}lcccc}
        \toprule
        \multirow{2}[2]{*}{\textbf{{Method}}} & \multirow{2}[2]{*}{\textbf{\makecell{Additional\\Data Type}}}& \multicolumn{3}{c}{\textbf{IF}} \\
        \cmidrule{3-5}
        & & \textbf{100} & \textbf{50} & \textbf{10}\\
        \midrule
        CE$_\texttt{ CVPR'19}$  & N/A & 37.96 & 43.54 & 59.50 \\ 
        Mixup$_\texttt{ ICLR'18}$  & N/A & 36.21 & 41.09 & 56.82 \\
        Cutmix$_\texttt{ ICCV'19}$ & N/A & 37.91 & 43.54 & 60.27 \\
        \cmidrule{1-5}
        Intra-class Image Translation & Syn. & 47.87 & 53.33 & 64.95 \\
        Inter-class Image Translation & Syn. & 47.17 & 51.33 & 64.11 \\
        Class Distribution Fitting     & Syn. & 51.53 & 55.60 & 65.60 \\
        Web crawled images    & Real & 54.06 & 56.40 & 63.86 \\
        \cmidrule{1-5}
        SYNAuG                          & Syn. & \textbf{58.59} & \textbf{61.36} & \textbf{69.01} \\
        \bottomrule 
    \end{tabular}
    }
\end{table}

\noindent\textbf{How good is SYNAuG for generating samples? Much better than web crawled images.}
We present other variants of generation methods as baselines:
1) Motivated by the recent work~\citep{he2022synthetic} using the few-shot original samples as guidance during the generation process, we first introduce \emph{Intra-class Image Translation}, where we use the original samples from the original training data as a class-wise guidance image for generation, 
2) Inspired by the M2m~\citep{kim2020m2m} translating an image of the major class to the minor class for leveraging the diversity of the majority information, we introduce \emph{Inter-class Image Translation}, where we use random samples in the dataset as guidance regardless of the class, 
3) As an advanced version motivated by DreamBooth~\citep{ruiz2022dreambooth}, we fine-tune the diffusion model with the samples in each class to model the class-wise distribution, named \emph{Class Distribution Fitting},
and 4) As a strong baseline, we collect the real data from the internet instead of generating synthetic images, \ie, \emph{Web crawled images}.

In \Tref{tab:cifar100_lt_baseline}, we compare our method with proposed baselines.
Compared to the case that uses real-world web data, it shows that the generated images are of sufficient quality to mitigate the class imbalance problem.
Also, we evaluate additional baselines, which apply the variant methods during the generation process.
While they are better than training only with the original long-tailed data (CE method), the performance is lower than SYNAuG.
While simple, SYNAuG is noticeably effective.

\noindent\textbf{Is SYNAuG compatible with other algorithms to make a synergy? Yes!}
In \Tref{tab:fairness_2}, we evaluate whether our method can be compatible with two algorithms,
Group-DRO~\citep{sagawa2019distributionally} and Re-Sampling (RS).
Note that we do not apply Mixup and fine-tuning in this experiment.
Group-DRO and RS improve the fairness metrics of ERM at the same time.
While Group-DRO and RS improve SYNAuG in some fairness metrics, developing a fairness algorithm with synthetic data might be a promising direction toward a fair model.

In \Tref{tab:robustness}, SYNAuG generates samples not to be correlated with spurious features, which improves the performance in BaseModel both on worst and mean accuracies.
When applying DFR, the synthetic data can increase the worst and mean accuracy consistently.
We also observe that fine-tuning is more effective compared to re-train, which is consistent with \Tref{tab:ablation}.

\begin{figure}[t]
    \centering
    \includegraphics[width=0.95\linewidth]{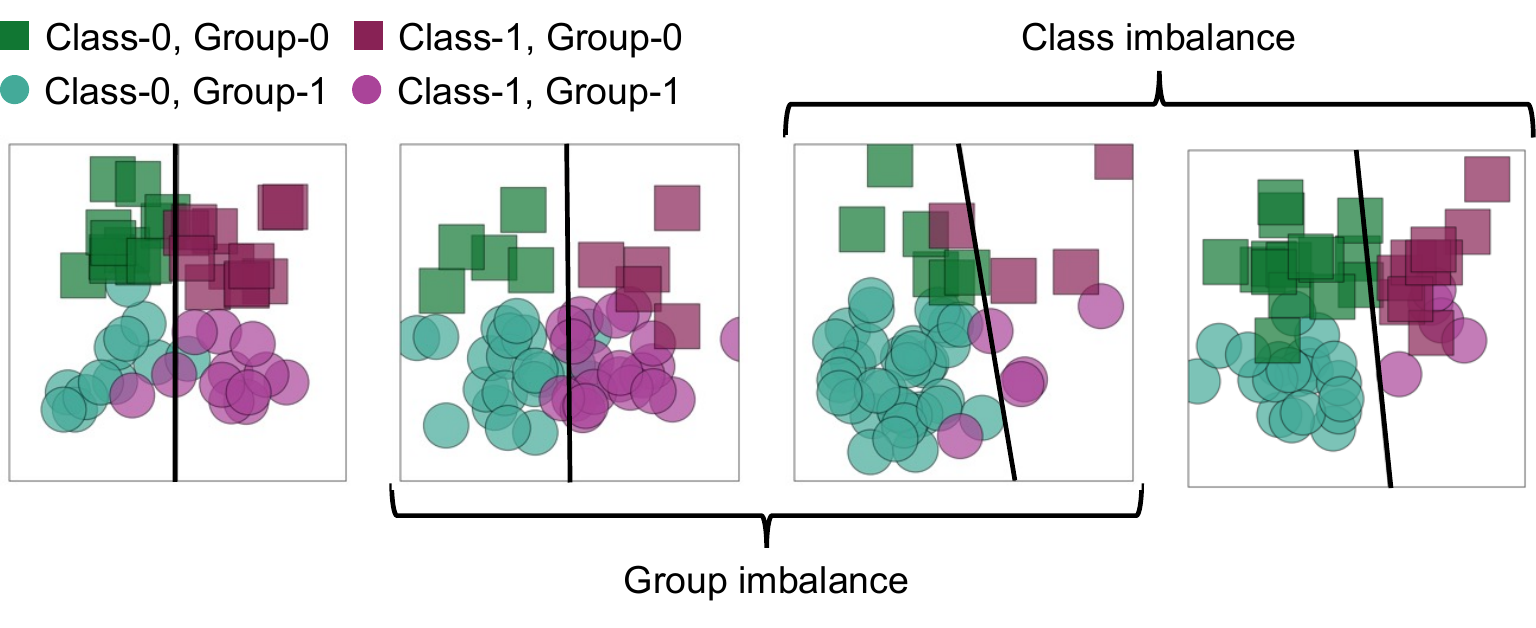}
    \caption{\textbf{Influence of the class and group imbalance on classifier.
    } 
    The 2D data are sampled from the normal distributions with four different means and the same covariance. 
    We simulate 4 different experiments with the latent group imbalance (sensitive attributes) by adjusting the number of data in each group.
    We train classifiers for the classes and visualize the learned classifiers (bold black lines).
    The fairer the classifiers, the more vertically aligned.
    The classifier trained on the class imbalance is more unfair than the one on the group imbalance.
    }
    \label{fig:fair_toy}
\end{figure}

\noindent\textbf{Can we improve fairness performance even when we do not know about sensitive attributes? Favorably, yes.}
In \Fref{fig:fair_toy}, we empirically observe that the group imbalance with the class imbalance amplifies unfair classifiers.
While both class and group imbalance contribute to the unfair classifiers, class imbalance contributes more than group imbalance.

In \Tref{tab:fairness_4}, we compare the performance between original SYNAuG and variant SYNAuG$^*$.
For SYNAuG, we augment the data to mitigate the class imbalance across the sensitive attribute; the female and male ratio of each sensitive attribute becomes equal.
For SYNAuG$^*$, we augment the synthetic data to mitigate the class imbalance regardless of sensitive attributes.
Although exploiting the knowledge of sensitive attributes is more effective, SYNAuG$^*$ shows better fairness metrics compared to ERM.

\section{Conclusion}
We propose SYNAuG to investigate the effectiveness of synthetic data in data imbalance problems.
The key findings are the importance of at least a few real-world samples and the existence of a domain gap for synthetic images from a recent diffusion model.
Despite the above limitations of synthetic data, SYNAuG is effective on data imbalance problems, including long-tailed recognition, model fairness, and robustness to spurious correlations.
Our study suggests the importance of controlling imbalance from the data perspective.
We believe that addressing data controllability is a promising research direction to resolve the early bottleneck in machine learning model development. 
While we focus on the data perspective, improving the model in multiple views is necessary to solve data imbalance effectively.

\bibliography{refs}
\bibliographystyle{icml2023}



\end{document}